\documentclass{bmvc2k}

\usepackage{booktabs}
\usepackage{amssymb}
\usepackage{amsmath}
\usepackage{subcaption}

\newcommand{\ours}{STS\xspace}

\title{Dense Video Captioning using State-Space Models with Transfer State}
\title{Time-Scaling State-Space Models for Dense Video Captioning}

\addauthor{AJ Piergiovanni}{ajpiergi@google.com}{1}
\addauthor{Ganesh Satish Mallya}{ganeshmallya@google.com}{1}
\addauthor{Dahun Kim}{mcahny@google.com}{1}
\addauthor{Anelia Angelova}{}{1}

\addinstitution{
 Google Deepmind}

\runninghead{Piergiovanni, Mallya, Kim, Angelova}{Time-Scaling State-Space Models}

\begin{document}

\maketitle

\begin{abstract}
Dense video captioning is a challenging video understanding task which aims to simultaneously segment the video into a sequence of meaningful consecutive events and to generate detailed captions to accurately describe each event. 
 Existing methods often encounter difficulties when working with the long videos associated with dense video captioning, due to the computational complexity and memory limitations.
 Furthermore, traditional approaches require the entire video as input, in order to produce an answer, which precludes online processing of the video.
 We address these challenges by time-scaling State-Space Models (SSMs) to even longer sequences than before. Our approach, State-Space Models with Transfer State, combines both the long-sequence and recurrent properties of SSMs and addresses the main limitation of SSMs which are otherwise not able to sustain their state for very long contexts, effectively scaling SSMs further in time.
 The proposed model is particularly suitable for generating captions on-the-fly, in an online or streaming manner, without having to wait for the full video to be processed, which is more beneficial in practice.
 When applied to dense video captioning, our approach scales well with video lengths and uses 7x fewer FLOPs.
\end{abstract}


\section{Introduction}
\vspace{-0.2cm}


Dense video captioning is a challenging task which involves accurately localizing 
multiple events by specifying their start and end timestamps, and describing each event with a detailed caption. It requires understanding of the video in depth, both at the local level and grasping the content across longer time horizons. 
Dense video captioning is typically conducted over long videos~\cite{youcook2,vitt,ActivityNet} 
that contain multiple diverse activities. 
It requires efficient processing of a large number of frames and modeling of their long-range dependencies. 



\begin{figure}[t]
\begin{center}
\includegraphics[width=0.77\linewidth]{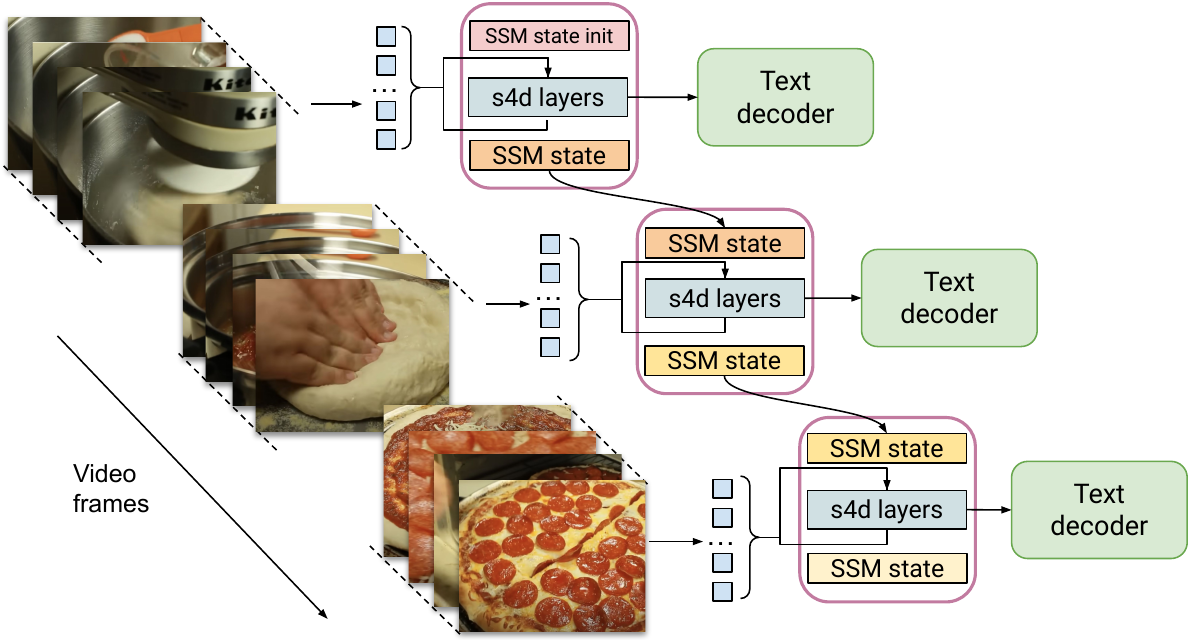}
\end{center}
\caption{
We propose State-Space Models with Transfer State to extend 
SSMs to tasks such as dense video captioning that require longer video inputs. Leveraging the linear scaling properties of SSMs and combining that with their inherent statefulness, we maintain the SSM state over consecutive SSM modules and update and propagate it iteratively across the whole video. This is much more cost-efficient than processing the full video at the same time. 
The dense captioning and localization outputs are produced at various time-steps by local decoders, as the video is processed. 
Each decoder possesses the full state of context prior to that point, which is important for generating context-aware and accurate captions.
This enables dense captioning to be done on-the-fly, without loading the full video in memory. 
}
\label{fig:teaser}
\end{figure}

State-Space Models (SSMs) provide a powerful framework for dynamic, sequential modeling making them particularly suitable for analyzing videos. 
SSMs have been recently proposed~\cite{SSM,s4d,s4nd,Mamba} as an alternative to the Transformer architectures~\cite{vaswani2017attention} which can be expensive to compute for longer input signals due to the quadratic scaling of attention mechanisms with increasing sequence lengths. SSMs~\cite{SSM} have efficient, near linear-scaling processing of the sequence, which holds promise for longer video understanding. Many recent works
apply SSM modeling to videos~\cite{VideoMamba,VideoMambaPark,VideoMambaSuite,Wang2023s4longvideo,Islam2022LongMovie,Islam2023EfficientMovie}, in some cases for long video understanding tasks~\cite{VideoMambaSuite,Wang2023s4longvideo,Islam2022LongMovie,Islam2023EfficientMovie}. 
While SSMs allow better scaling to longer sequences, the effective video length considered in the above-mentioned works~\cite{VideoMamba,VideoMambaPark,VideoMambaSuite,Wang2023s4longvideo,Islam2022LongMovie,Islam2023EfficientMovie} is relatively small, i.e., as Table~\ref{tab:stats} shows, at most 64 frames~\cite{VideoMamba} are used. Even long videos processed by SSMs have not exceeded 60 seconds~\cite{Islam2022LongMovie}. 
This is because it is still not trivial to scale SSMs to long videos, because each video frame already has a large number of visual tokens, taking a lot  of memory. At the same time, current SSMs require all their inputs to be loaded in memory~\cite{VideoMamba,VideoMambaPark,VideoMambaSuite,Wang2023s4longvideo,Islam2022LongMovie,Islam2023EfficientMovie}, which
causes computation issues for longer sequences and severely inhibits the model from scaling to longer videos. 

On the other hand, SSMs can be computed recurrently and are in fact `stateful', i.e., have an internal state. However, the state starts to deteriorate with longer input lengths or for dynamic changes in the signal, as shown in~\cite{Illusion2024}. Thus its main stateful advantage is diminished in practice, 
and is less helpful for complex tasks such as dense video captioning and event localization, which require understanding of dynamic content over long videos. 

To that end, we propose a State-Space Model with a Transfer State (\ours, for short) to capture the evolving video content and generate captions that are both semantically meaningful and temporally aligned and to seamlessly scale to arbitrarily long videos. 
The key part of the design is to maintain and propagate the Transfer State across the segments of the video, specifically processing local video information sequentially and maintaining, reusing, and propagating the SSM Transfer State iteratively across longer video sequences (Fig.~\ref{fig:teaser}). 
With \ours, we unify the linear-scaling processing properties of SSMs with the stateful advantages of the recurrent version.
While fully leveraging the properties of SSMs, our approach neither processes the input one-step-at-a-time updating the state, nor requires the full set of tokens or video inputs to be present, as the proposed transfer state can be carried over for very long inputs. 
This allows the model to efficiently handle long video inputs while preserving temporal context. 
The approach is specifically suitable to the online settings and for challenging long video tasks e.g., dense video captioning. 
It allows the SSM approach to scale to arbitrarily long videos, without requiring the full video input to be present at all times. 



Prior methods address some of the challenges associated with long videos by hierarchical SSM methods~\cite{Bhirangi2024Hier} or by progressively pooling of information~\cite{Islam2022LongMovie,Islam2023EfficientMovie}, which still requires the full video input. This further limits the scalability, as these methods do not have mechanisms to preserve video information if the video cannot be fully loaded in memory.

We demonstrate strong performance on the difficult task of dense video captioning and event localization. 
Our approach allows both 1) time-scaling of SSM-based video models with respect to the video length and 2) applying them to online video understanding settings and for long videos.
The experiments show consistent gains in increasing the number of frames in the video and we observe significant benefits in processing the videos with online SSM with Transfer State.
Our model uses \textbf{7x fewer FLOPs}, compared to prior works.


The contributions of this work are:
\vspace{-0.3cm}
\begin{itemize}
    \item We propose State-Space models with Transfer State (\ours) where the state is maintained, reused and propagated across longer input sequences. This combines the efficiency of SSMs and their stateful properties and extends to much longer sequences.  
    \vspace{-0.7cm}
    \item We present efficient update rules for the state itself, and prove the equivalence of the Transfer State, verifying that the state will be preserved over long sequences. 
    \vspace{-0.3cm}
    \item We apply this to online dense video captioning, where both captioning and event localization can be obtained on-the-fly and without needing the full video to be loaded in memory. Interestingly, the proposed \ours encourages the SSM to sustain more information in the last snippet's tokens. This is in contrast with the standard SSM model for which we observe that this capability diminishes for longer sequences.   
\end{itemize}

\vspace{-0.4cm}
\section{Previous work}
\vspace{-0.2cm}


Many video understanding models~\cite{wang2022git,merlot,arnab2021vivit,MaMMUT,dynpretr,palix,flamingo} employ the Transformer architecture~\cite{vaswani2017attention}, adapted to vision tasks 
\cite{mvit,videoswin,piergiovanni2022tubevit,timesformer}.
However, due to the quadratic scaling of attention mechanisms,  the computational cost of Transformers is very high. This problem is exacerbated for videos, especially for long videos, as they require many more visual features. 

Recent works have proposed State-Space Models (SSMs), e.g.,~\cite{SSM}, Structured SSMs S4~\cite{s4}, S4D~\cite{s4d}, S4ND~\cite{s4nd}, S5~\cite{s5}, and Mamba~\cite{Mamba} for efficient processing of long sequences.
Following initial successful demonstration of SSMs for long-range tasks~\cite{SSM,s4,Mamba},
SSMs have been adapted to vision tasks~\cite{event_cameras,VisionMamba,Malik2024Towards,Islam2022LongMovie,SSMVideoDiffusion,Islam2023EfficientMovie}. 
SSMs for videos, are in their nascent stages.
In an early demonstration by Gu et al.~\cite{s4nd} SSMs are successfully applied to videos by swapping the I3D kernels~\cite{Carreira_2017_CVPR} of the ConvNeXt backbone~\cite{ConvNeXt} with SSM ones. 
Most prior video SSM models treat video as rasterized image sequences in time, which are hard to scale, due to the vast number of tokens from densely sampling of the videos.
 As a result, they consider relatively small video inputs e.g., 8 to 64 frames at most~\cite{s4nd,VideoMamba,VideoMambaPark,Islam2023EfficientMovie}, as seen in  Table~\ref{tab:stats}. 
Alternatively, the videos are processed with SSMs or Transformers which are then combined with a second-level of SSM models~\cite{Bhirangi2024Hier,Islam2022LongMovie,Islam2023EfficientMovie}.
These methods also work on relatively short videos and require the information from the full sequence. 



Dense video captioning~\cite{DensecapMasked,vid2seq,end2enddense,e2edensecap,zala2023hierarchical,iashin2020abetter} is a challenging task, conducted over untrimmed videos of several minutes, e.g., 2-5,~\cite{ActivityNet,vitt,youcook2}, sampling so few frames is not sufficient for providing dense and detailed video captioning per videos. With the exception of
~\cite{VideoMambaSuite}, which benchmarked various video tasks with the Mamba architecture, we are not aware of SSM methods applied to dense video captioning. Our work proposes a new mechanism which can leverage the SSM state and scale SSMs to longer videos beyond 64 frames. 
Some earlier dense video captioning approaches first detect the events boundaries, and then caption each event in a two-stage manner~\cite{iashin2020abetter}. Other methods, e.g.,~\cite{wang2018bidir,end2enddense,zhang2022unifying,zala2023hierarchical,end2enddense} unify the prediction of these jointly. 
Zhu et al.~\cite{e2edensecap} and  Yang et al.~\cite{vid2seq} tokenize the outputs for localization and captioning, formulating both tasks as sequence-to-sequence generation. 

The majority of the above-mentioned dense captioning methods are global, 
i.e., they are only able to produce an output after the full video is processed.
Our model works online instead, and does not require the full video or future video frames. 
Online methods have been rare, with only very recent work~\cite{zhou2024streaming} proposing token accumulation and clustering-based reduction. 
Our model proposes an alternative design where dense captioning are provided at regular intervals of the video and SSM propagates its state to arbitrarily long videos. 




\begin{table} [] 
\small
\centering
\begin{tabular}{l|cc}
Approach & Video length (sec.)	& Max num. frames   \\ 
\midrule
Gu et al.~\cite{s4nd} &2 &30  \\
VideoMamba, Park et al.~\cite{VideoMambaPark} &10 &32   \\ 
VideoMamba, Li et al.~\cite{VideoMamba} &10 &64    \\  
ViS4mer~\cite{Islam2022LongMovie} &60 &60    \\
\midrule
\ours (Ours) &120-315  &256  \\
\bottomrule
\end{tabular}
\vspace{3mm}
\caption{Recent SSM-based video approaches reach only 60 seconds video lengths and up to 64 frames. Dense video captioning requires much longer videos and more frames.}
\label{tab:stats}
\vspace{-0.3cm}
\end{table}







\begin{figure*}[t]
\begin{center}
\includegraphics[width=.77\linewidth]{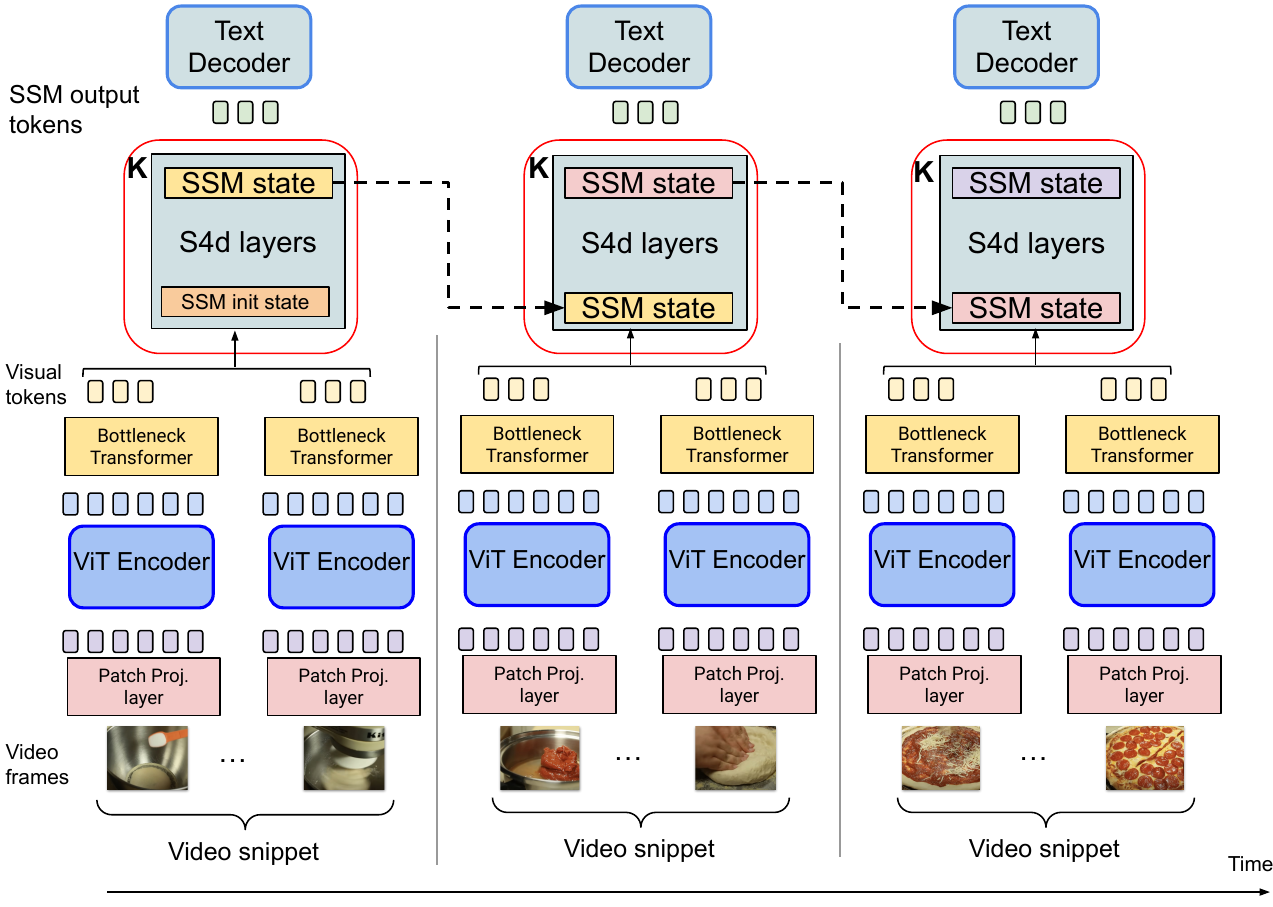}
\end{center}
\vspace{-3mm}
\caption{
At a high level, 
visual tokens from frames, belonging to video snippets, are fed to 
a sequence of SSM modules. 
The proposed SSM with Transfer State method computes and propagates the state between them, allowing the model to preserve the actual SSM state from previous inputs and scale to very long sequences. Note that the earlier parts of the video inputs are no longer needed after the state is computed. This facilitates efficient understanding for long videos, modeling both local frame features while preserving long-range dependencies across the video. Furthermore, it allows for online caption generation and processing, in theory, of infinite video by propagating only the state from prior video history.  
 }
\label{fig:main_model}
\end{figure*}

\vspace{-0.3cm}
\section{Approach}
\vspace{-0.2cm}
\noindent\textbf{SSM and S4D preliminaries:}
The State-Space Models (SSMs) are originally designed to model dynamic systems and have been shown to be well suited to visual or other real-world data~\cite{SSM}. State-Space models maintain a state internally which can track the evolution of these systems. 
%
%
%
An SSM in continuous time is represented by the Eq. \ref{ssm-hidden-ct} and \ref{ssm-output-ct} in continuous time and by Eq. \ref{ssm-hidden-dt} and \ref{ssm-output-dt} when discretized (here $A,B,C,D,\overline{A},\overline{B},\overline{C},\overline{D}$ are learnable tensors):
\begin{equation}
\label{ssm-hidden-ct}
    h'(t) = Ah(t) + Bx(t)
\end{equation}
\vspace{-5mm}
\begin{equation}
\label{ssm-output-ct}
    y(t) = Ch(t) + Dx(t)
\end{equation}
\vspace{-5mm}
\begin{equation}
\label{ssm-hidden-dt}
h = \overline{A}h + \overline{B}x
\end{equation}
\vspace{-5mm}
\begin{equation}
\label{ssm-output-dt}
    y = \overline{C}h + \overline{D}x,
\end{equation}
\noindent where $x(t)$ is the time-dependent input, $y(t)$ is the output, $h(t)$ is the hidden state and $t$ denotes time. When discretized, we use vectors $x$, $y$ and $h$ to represent a sequence of the discretized values for the inputs, outputs and state, respectively. 
During training of the SSMs, since the data is available beforehand, a convolution with a specifically designed kernel, as shown in the S4D SSM variant~\cite{s4d}, can be employed for efficient training, over the whole input sequence. This is shown in Eq. \ref{ssm-output-conv} using the shortcut notation $*$
for convolution: 
\begin{equation}
\label{ssm-output-conv}
    y = \overline{K} * x,
\end{equation}
\noindent where $\overline{K} = [\overline{CB}, \overline{CAB}, ..., \overline{CA^{(L - 1)}B}]$. 
Here, as seen, the output is produced in a single compute operation (i.e.,  a single convolution) where the convolution kernel $\overline{K}$ has a special structure~\cite{s4d}. 
The advantage of this derivation is that convolutions can be implemented with Fast Fourier Transformations (FFT) with an $n\log(n)$ complexity for a sequence length of $n$, giving an edge to the quadratic ($n^2$) scaling of Transformers. 
This approach requires the full input sequence which can be a drawback if the sequence is very large, as the corresponding convolution kernel will be large too, which in turn causes both memory and efficiency issues.

\vspace{-0.3cm}
\subsection{SSMs with Transfer State (\ours)} 
\label{sec:ssm_transfer}
\vspace{-0.2cm}
As shown in Eq. \ref{ssm-output-conv}, the convolutional formulation requires the full input sequence to be available, and this limits the applicability to inputs that can fully fit into memory. Despite the better scaling of SSMs to long sequences, this will still be a problem when handling long videos, due to the corresponding abundant video features. An alternative view is the purely recurrent view of SSMs where the output is computed at each step given the input and hidden state (i.e., Eq. \ref{ssm-hidden-ct} and \ref{ssm-output-ct}). This dependency causes inefficiency in computation, where each value needs to be computed before the next one.
We further observe from Eq. \ref{ssm-output-conv} that the state $h$ is no longer directly computed. However, the state represents the underlying semantics and temporal context of the video, and has distilled important information from the input. 

To scale SSMs to arbitrarily long videos, we propose a State-Space Model with a Transfer State (\ours) to continuously capture the evolving content of video (Figure~\ref{fig:teaser}). Here, the idea is that the Transfer State is computed over portions of the input and the model propagates its Transfer State to subsequent SSM computations across the video. This allows loading and processing shorter and manageable video portions at a time, while still maintaining the SSM state to handle the full length video without losing any information. More specifically, the model processes a part of the video with SSM, using the efficient computation from Eq.~\ref{ssm-output-conv}, but is also explicitly computing and returning the Transfer State value for future reuse. This returned state is used as the initial state when processing the next video snippet, thus computing the same values as if the whole sequence were given at once, preserving the information of the entire context, but not requiring the entire input sequence to be loaded in memory. 

\vspace{-0.2cm}
\subsection{Per-segment state update}
\vspace{-0.1cm}
However, in order to propagate the Transfer State, we need to compute the state explicitly at a segment level. In the efficient convolutional derivation, the state is not explicitly computed, i.e., Equation \ref{ssm-output-conv}, and similarly the recurrent version does not yield the state without having to compute it step-by-step.  To address this, we observe that any (hidden) state for $L$ timesteps is also derived via convolution where the kernel for convolution is as shown below in Eq. \ref{hidden-state-kernel}.
\begin{equation}
    \label{hidden-state-kernel}
    \overline{K_{hidden}} = [\overline{B}, \overline{AB}, ..., \overline{A^{(L - 1)}B}].
\end{equation}
For subsequent video snippets, the hidden state from the previous snippet is fed in (say ${h_0}$) thereby allowing the model to retain long range information from previous snippets. The output and the hidden states when given a previous hidden state are computed as in Eq. \ref{output-with-prev-hidden}, \ref{hidden-with-prev-hidden}.
\begin{equation}
    \label{output-with-prev-hidden}
    y = \overline{K} * x + \overline{CA^{(L - 1)}}h_0
\end{equation}
\vspace{-5mm}
\begin{equation}
    \label{hidden-with-prev-hidden}
    h = \overline{K_{hidden}} * x + \overline{A^{(L - 1)}}h_0.
\end{equation}
This `divide-and-conquer' style approach avoids both the need to handle excessively long videos, provides efficient computation of the activations as well as the hidden states. It further allows for online processing of the video and the output can be produced at multiple locations alongside the video and at any point where activations and state are available.
Our approach is different from other SSM approaches which also subdivided the input sequence, e.g., hierarchical SSMs~\cite{Islam2023EfficientMovie,2023videollm}, in two main respects: 1) we maintain and propagate the state which is the de-facto internal representation of the sequential SSM model, whereas the state is ignored in the prior approaches as they work on accumulating activations 2) the full input sequence (i.e., the full set of features from the whole video) is still required for the hierarchical computation, thus they are not easily applicable to the online settings, and many sets of activations need to be stored.

\vspace{-0.2cm}
\subsection{Transfer State equivalence}
\vspace{-0.1cm}
\label{sec:equiv}

One main question is whether the proposed SSM with Transfer State (\ours) is equivalent to the full SSM computation, i.e., if we were to be able to access the full video. 
We show that propagating the state in STS is equivalent to the state in the original SSM. The \ours however provides advantages, as the whole input, and activations, do not need to be fully loaded in memory and can be processed online and discarded once used. This is in contrast to prior works, e.g. hierarchical SSMs and the standard, full SSM, which need the full set of inputs.
Given an input $\{x\}$ of sequence length $L$, splitting the signal into two and feeding the first $M$ values of the signal into SSM will result in the following, where $k \in [1, L]$
\begin{equation}
    h_k = \overline{A} * h_{k - 1} + \overline{B} * x_k
\end{equation}
\vspace{-5mm}
\begin{equation}
    y_k = \overline{C} * h_k + D * x_k
\end{equation}
Unrolling the above recursion we get:
\begin{equation}
    \label{hidden-state-first-chunk}
    h_{M} = \overline{A}^{M - 1}\overline{B} * x_1 + \overline{A}^{M - 2}\overline{B} * x_2 + ... + \overline{B} * x_{M}.
\end{equation}
Feeding the hidden state from the first $M$ values (Eq. \ref{hidden-state-first-chunk}) to the SSM when feeding the rest of the $L - M$ signal values will result in the following, after unrolling the recursion.
\begin{align}
    \label{hidden-state-from-chunking}
    {h_{L}} = \overline{B} * {x_{L}} + \overline{A}\overline{B} * {x_{L - 1}} +   \overline{A}^{L - M - 1}\overline{B} * x_{L - M + 1}  \nonumber 
    + \overline{A}^{L - M} * h_{M} \nonumber \\
    h_{L} = \overline{B} * {x_{L}} + \overline{A}\overline{B} * {x_{L - 1}} + ... + \overline{A}^{L - M - 1}\overline{B} * x_{L - M + 1} \nonumber \\
    + \overline{A}^{L - M} * (\overline{A}^{M - 1}\overline{B} * x1 + ... + \overline{B} * x_{M}) \nonumber \\
    h_{L} = \overline{B} * {x_{L}} + \overline{A}\overline{B} * {x_{L - 1}} + ... + \overline{A}^{L - M - 1}\overline{B} * x_{L - M + 1} \nonumber \\
    + \overline{A}^{L - M}\overline{B} * x_{L - M + 1} + ... + \overline{A}^{L}\overline{B} * {x_1}.
\end{align}
Eq. \ref{hidden-state-from-chunking} shows that the final hidden state produced by segmenting and splitting the input signal is the same as providing the entire input signal to the SSM, which proves the equivalence.






\vspace{-0.2cm}
\subsection{Application to online dense video captioning}
\vspace{-0.1cm}

Visual tokens are first extracted from the frames of consecutive video snippets using a ViT encoder~\cite{dosovitskiy2020image} with a standard $16\times 16$ patch projection. We  then reduce the number of tokens from each frame using a 'bottleneck' Transformer~\cite{mirasol3B} by taking only the last desired $N$ number of tokens, i.e., a video snippet might have $T$ frames they will generate $N*T$ tokens. 
The visual tokens are fed to a sequence to SSM modules, which sequentially process their portion of the visual tokens, compute the SSM state (the Transfer State) and then propagate the Transfer State to the subsequent SSM module  (Figure~\ref{fig:main_model}). Each SSM initializes its state from the previous Transfer state preserving the overall SSM state. This, as shown in Eq.~\ref{hidden-state-from-chunking} is equivalent to the SSM processing the full input sequence, but has the advantage that the computation is split and information can be discarded once the state is computed.
This enables the SSM model to retain its state across multiple video snippets, allowing the model to scale to long videos, while still modeling the temporal information in the video. We use local text decoders, similarly to the global one in \cite{vid2seq}, but focusing on shorter snippets.

Due to this design, the model can handle longer sequences. 
For example, taking 1024 frames (256 tokens per frame) and splitting them in segments of 16 frames, this gives 64 consecutive segments, each segment of 4096 tokens.
Each segment can easily fit into memory and be processed by the model. In contrast, a Transformer-based model taking all the tokens at once would need to handle 262k tokens, i.e., significantly more. Further, if we applied a Transformer per segment, similarly to the SSM, the Transformer does not have the state to propagate to the next segment, which makes it hard to model long-term temporal structure. 
%
%


\textbf{Implementation details:}
The SSM model has 46M parameters, 
the ViT-L has 300M, the shared decoder has 128M and the bottleneck transformer has 128M as well. Overall the model has about 600M parameters. 
 We use an image resolution of 352 and 8 segments per video and 128 frames per video for most experiments, but show increasing to 256 frames has additional benefits.  In the decoders, we encode the time by discretizing the timestamps into 32 buckets after normalizing the time based on video length. 

\vspace{-0.3cm}
\section{Experimental results}
\vspace{-0.2cm}



We conduct experiments on three popular dense video captioning benchmarks: \textbf{VITT~\cite{vitt}}, \textbf{ActivityNet~\cite{ActivityNet}}, and \textbf{YouCook2~\cite{youcook2}}, reporting results using the previously established metrics and protocols.
METEOR \cite{banerjee2005meteor}, CIDEr \cite{vedantam2015cider}, and SODA \cite{fujita2020soda} metrics are reported for dense captioning and F1 score  which evaluates the precision of temporal localization of segments, based on IoU with the ground truth event boundaries. We report IoU in ablations.

\begin{table*}[t]
\centering
\small
\begin{tabular}{l|cccc|cccc|cccc}
& 
  \multicolumn{4}{c|}{ViTT} & \multicolumn{4}{c|}{YouCook2} & \multicolumn{4}{c}{ActivityNet} \\
 Method  & S    & C     & M    & F1     & S   & C   & M   & F1     & S   & C   & M   & F1   \\
\midrule

UEDVC\scriptsize{\citep{zhang2022unifying}}  
     & - & - & - & -     & 3.3  & 8.4  & 2.2 & -    & 5.5 & - & - & -    \\
Masked\scriptsize{\citep{DensecapMasked}}  
  & - & - & - & -     & - & 6.1 & 3.2 & -    & - & 9.3 & 5.0 & -    \\
PDVC\scriptsize{\citep{end2enddense}}  
    & - & - & - & -     & 4.9 & 28.9 & 5.7 & -    & 6.0 & 29.3 & 7.6 & -    \\
E2ESG\scriptsize{\citep{e2edensecap}}  
    & - & - & - & -     & - & 25.0 & 3.5 & -    & - & - & - & -    \\    
TimeCh\scriptsize{\citep{ren2024timechat}}  
     & - & - & - & -      & 3.4  & 11.0 & - & 19.5    & - & - & - & -  \\

OmniV\scriptsize{\citep{wang2024omnivid}}  
     & - & - & - & -       & - & - & - & -     & - & 26.0 & 7.5 & - \\

Vid2Seq\scriptsize{\citep{vid2seq}} 
    & 9.8 & 23.0 & 5.0 & \textbf{37.7}      & 5.7 & 25.3 & 6.4 & 23.5    & 5.9 & 30.2 & 8.5 & 51.8    \\

\midrule

DIBS$\dagger$\scriptsize{\citep{wu2024dibs}} 
  & - & - & - & -      & 6.4 & \textbf{44.4} & 7.5 & 31.4    & 5.9 & 31.9 & 8.9 & \textbf{55.6}    \\
%

SDC$\dagger$\scriptsize{\cite{zhou2024streaming}} 
    & \textbf{10.0} & 25.2 & 5.8 & 35.4      & 6.0 & 32.9 & 7.1 & 24.1    & 6.2 & \textbf{37.8} & 10.0 & 52.9    \\

GIT$\dagger$\scriptsize{\citep{wang2022git}}  
      & 7.1 & 15.1 & 3.4 & 32.5     & 3.1 & 12.1 & 3.4 & 17.7    & 5.7 & 29.8 & 7.8 & 50.6    \\

\midrule 

\small{VMS\scriptsize{~\cite{VideoMambaSuite}}} 
    & - & - & - & -
    &4.3  &22.1  &4.4  & 28.4 
    &5.3  &26.8  &7.2  & 54.3 \\

\textbf{\ours Ours} 
   &8.5 &\textbf{42.4} &\textbf{19.6} & 34.9    
   & \textbf{9.7} & 37.8  & \textbf{18.7}  & \textbf{37.7}   
   & \textbf{8.9} & 33.4 & \textbf{16.0} &  41.6 \\
\bottomrule
\end{tabular}

\vspace{3mm}
\caption{
\textbf{Dense Video Captioning and Event Localization results compared to SOTA}.  VITT, YouCook2 and ActivityNet datasets.  We use the standard evaluation metrics: SODA (S), CIDEr (C), METEOR (M), and localization (F1 score).
Our method does not use video pre-training, similar to the models at top. The models in the middle section use large pre-training (e.g. YT-Temporal-1B~\cite{vid2seq}, HowTo100M~\cite{wu2024dibs}, WebLI~\cite{zhou2024streaming}).  
The bottom section, including ours, are SSM-based approaches, all others use Transformers.
Vid2Seq is the version with visual-only inputs. 
$\dagger$: Methods which use datasets that are not public.
}
\label{tab:sota}
\vspace{-5mm}
\end{table*}


\noindent \textbf{Comparison to SOTA.} Table~\ref{tab:sota} shows the main results of our (\ours) approach compared to the state-of-the-art (SOTA).
%
Our model shows strong performance, outperforming most prior SOTA approaches, despite not using additional video pre-training.
We note that DIBS~\cite{wu2024dibs} and SDC~\cite{zhou2024streaming} both use large and non-public video pre-training datasets. 
Our model makes large gains in SODA and METEOR  \textbf{(1.5x)} for YouCook2 and ActivityNet and CIDEr and METEOR for VITT.
Our model outperforms the other SSM-based approach, VideoMambaSuite (VMS)~\cite{VideoMambaSuite} by large margins.
We compare to video-only inputs as this is adopted in most contemporary dense video captioning evaluation protocols, and because the additional ASR input, used in early works~\cite{vid2seq}, artificially boosts performance, as it is often aligned with captions. 
Our vision-only setting is more challenging, relying on only video inputs.

\noindent \textbf{Efficiency.} Our approach uses \textbf{7x fewer FLOPs} in comparison to prior work (Table~\ref{tab:flops}), e.g., by our calculation \cite{zhou2024streaming} uses 12800 GFLOPs for 128 output tokens. Our \ours uses only 1854 GFLOPs, or 231 GFLOPs per segment, which can be linearly scaled to any video length.


\begin{figure}
\begin{minipage}{0.7\textwidth}
\small
        \centering
        \begin{tabular}{ll|ccccc}
         &  & C & S & M & mIOU & F1 \\
        \midrule
        Global & Transformer &  15.0 & 5.1 & 6.3 & 38.1 & 27.4 \\
        Global & SSM & 14.7 & 5.1 & 6.2 & 40.3 & 29.1 \\
        \midrule
        Online & \ours (Ours) & \textbf{16.1} & \textbf{6.1} & \textbf{14.2} & \textbf{50.4} & \textbf{32.3}\\
      \bottomrule
      \end{tabular}
\vspace{3mm}
      \captionof{table}{Global single caption SSM vs. online SSM with Transfer State (\ours). A global counterpart using the standard Transformer is also shown. Uses 32 frames, resolution of 224, bottleneck size of 8, image resolution of 224 and state size of 8.
      }
      \label{tab:global-vs-local}
\end{minipage}\hfill%
\begin{minipage}{0.28\textwidth}
\centering
\small
\begin{tabular}{c|c}
\toprule
Model & GFLOPs \\
\midrule
  SDC \cite{zhou2024streaming}  & 12800  \\
  \ours (Ours)   & 1854 \\
\bottomrule
\end{tabular}
\vspace{3mm}
      \captionof{table}{Our model uses \textbf{7x fewer} FLOPs than competitive prior works.
      }
      \label{tab:flops}
\end{minipage}
\vspace{-3mm}
\end{figure}

\begin{table}
\small
    \centering
        \begin{tabular}{ll|ccccc}
        Method &Tokens  & C & S & M & mIOU & F1 \\
        \midrule
    SSM Global    &All Tokens (of sequence) &	14.1 &	5.2 &	6.1	& 38.3 & 27.4 \\
    SSM Global    &Last token per segment  & 14.7	 & 5.1 & 6.2 & 40.3 & 29.4 \\
    SSM Global    &Final Token (of sequence) & 13.5 & 5.0 & 6.3 & 39.0 & 30.3 \\
    \midrule
      \ours Online   &All Tokens (of sequence) & 14.7 & 5.8 & 11.0 & \textbf{52.2} & 30.1\\	
     \ours Online   &Last token per segment & \textbf{16.1}	& \textbf{6.1} &\textbf{14.2} & 50.1 & \textbf{32.4} \\	
        \bottomrule
       \end{tabular}
\vspace{3mm}
        \caption{Exploring the importance of which tokens are used for final caption generation. \ours applied online outperforms the global model as it 
        successfully accumulates information into the last token per segment. 
        Uses 32 frames, resolution of 224, state, bottleneck sizes 8.}
        \vspace{-0.3cm}
        \label{tab:which-tokens}
\end{table}


\noindent \textbf{Ablations.} Ablation experiments are done on the VITT dataset and, 
unless otherwise stated, use fewer frames (32) and smaller number of layers (8), or smaller resolution or dimensions, in order to save compute.
%
%
%
In Table \ref{tab:global-vs-local}, we compare the proposed online \ours to its global counterparts i.e., a Transformer or an SSM that process the entire video and then generate all the captions (similar to Vid2Seq~\cite{vid2seq}). 
\ours performs better than the global models using either a Transformer or SSM. 
%
In Table \ref{tab:which-tokens}, we study the importance of tokens in the sequence for \ours and a global captioning SSM. We note that while our proposed \ours is an SSM itself, its use in an online captioning setting provides advantages, as captions are generated when the actions occur, rather than all at the end. We find that when using the online \ours model, using the last token per segment is significantly better and outperforms even using all the tokens, suggesting that in this setting, each segment can be compressed to a single token representation, further facilitating online processing. This is notable because it performs better than using all the tokens and also reduces memory usage. In the final model, we used all the tokens, corresponding to the last frame of the segment, as input, as this gave a further small increase in performance.
When using the final token of the entire (global) sequence, the performance drops a lot, suggesting that the SSM is unable to represent the entire video in a single token, which is expected as the sequence is long. 
In Table~\ref{tab:num-frames} we see the scalability of our model with the number of frames. Increasing the number of frames consistently improves performance, all the way up to 256 frames. We also note that there is a nontrivial jump across all metrics, if we are to compare results for 128 or 256 frames, vs. e.g., 16-32.
Table~\ref{tab:transfer-state} compares the vanilla SSM vs. SSM with Transfer State (both online), where we expect these to match due to the preservation of the state, as shown in Sec.~\ref{sec:equiv}.

\section{Conclusions}
This paper proposes a novel approach for dense video captioning, State-Space Models with Transfer State, where the state is accumulated, updated and propagated across multiple input segments to handle any sizes of video. We present efficient update rules for the state itself.
Our model is particularly suitable for online video processing, generating dense video captions on-the-fly, without having to load the full video input in memory.
 


\begin{table}
 \centering
\small
        \begin{tabular}{c|ccccc}
       \# Frames & C & S & M & mIOU & F1 \\
       \midrule
        16 & 28.7 & 6.8 & 16.0 & 57.5 &  31.3 \\
        32 & 27.5 & 7.3 & 16.5 & 59.9 & 33.2 \\
        64 & 34.1 & \textbf{7.8} & 17.5 & 60.1 & 35.4 \\
        128 & 34.9 & \textbf{7.8} & 17.9 & 61.5 & 35.6 \\
        256 & \textbf{35.0} & \textbf{7.8} & \textbf{18.0} & \textbf{62.0} &\textbf{35.9} \\
        \bottomrule
       \end{tabular}
\vspace{3mm}
    \caption{Online \ours for varying frame numbers. Resolution 352, state, bottleneck sizes 16.}
    \label{tab:num-frames}
\end{table}

\begin{table}  [h]
\small
\vspace{-0.5cm}
\vspace{8mm}
    \centering
        \begin{tabular}{l|ccccc}
         & C & S & M & mIOU & F1 \\
        \midrule
        SSM & 23.8 & 5.7 & 9.6 & 39.6 & 27.8  \\
        SSM with Transfer State (\ours) & 24.0 & 5.7 & 9.6 & 39.7 & 27.6  \\
       \bottomrule
       \end{tabular}
\vspace{3mm}
     \caption{Studying \ours vs. standard SSM, online model. As expected they are comparable. 
     }
    \label{tab:transfer-state}
\end{table}

\pagebreak


\bibliography{egbib}
\end{document}